\pdfoutput=1

\documentclass[11pt]{article}

\usepackage[preprint]{acl}

\usepackage{times}
\usepackage{latexsym}

\usepackage[T1]{fontenc}

\usepackage[utf8]{inputenc}

\usepackage{microtype}

\usepackage{inconsolata}
\usepackage{xspace}

\usepackage{graphicx}

\usepackage{lipsum}  

\usepackage{booktabs}
\usepackage{xcolor}         
\usepackage{colortbl}
\usepackage{multirow}
\usepackage[normalem]{ulem}
\usepackage{hyperref}
\usepackage{enumitem}

\useunder{\uline}{\ul}{}

\usepackage{fancyhdr}
\definecolor{LightCyan}{rgb}{0.88,1,1}

\newcommand{\score}{\textsc{VideoScore}\xspace}
\newcommand{\data}{\textsc{VideoFeedback}\xspace}

\title{\score: Building Automatic Metrics to Simulate Fine-grained Human Feedback for Video Generation}

\author{
$^{1,2}$Xuan He$^*$$^\dagger$ \quad 
$^{1}$Dongfu Jiang$^*$$^\dagger$ \quad 
$^{1,3}$Ge Zhang \quad
$^{1}$Max Ku \\
$^{1}$\textbf{Achint Soni} \ \ 
$^{1}$\textbf{Sherman Siu} \ \ 
$^{1}$\textbf{Haonan Chen} \ \ 
$^{1}$\textbf{Abhranil Chandra} \ \ 
$^{1}$\textbf{Ziyan Jiang} \\
$^{1}$\textbf{Aaran Arulraj} \ \ 
$^{4}$\textbf{Kai Wang} \ \ 
$^{1}$\textbf{Quy Duc Do} \ \ 
$^{1}$\textbf{Yuansheng Ni} \ \ 
$^{2}$\textbf{Bohan Lyu} \\
$^{1}$\textbf{Yaswanth Narsupalli} \ 
$^{1}$\textbf{Rongqi Fan} \ 
$^{1}$\textbf{Zhiheng Lyu} \ 
$^{5}$\textbf{Bill Yuchen Lin} \ 
$^{1}$\textbf{Wenhu Chen }$^\dagger$ \vspace{0.4em} \\
$^1$University of Waterloo, $^2$Tsinghua University,
\\
$^3$Stardust.AI, $^4$University of Toronto, $^5$AI2 \vspace{0.4em}\\
$^*$ Equal Contribution \vspace{0.4em}\\
$^\dagger$ \texttt{\{x36he, dongfu.jiang, wenhuchen\}@uwaterloo.ca}
}

\begin{document}

\twocolumn[{%
    \renewcommand\twocolumn[1][]{#1}
    \maketitle
    \centering
    \vspace{2.5em}
    \url{https://tiger-ai-lab.github.io/VideoScore/}
    \begin{center}
        \centering
        \includegraphics[width=0.84\textwidth]{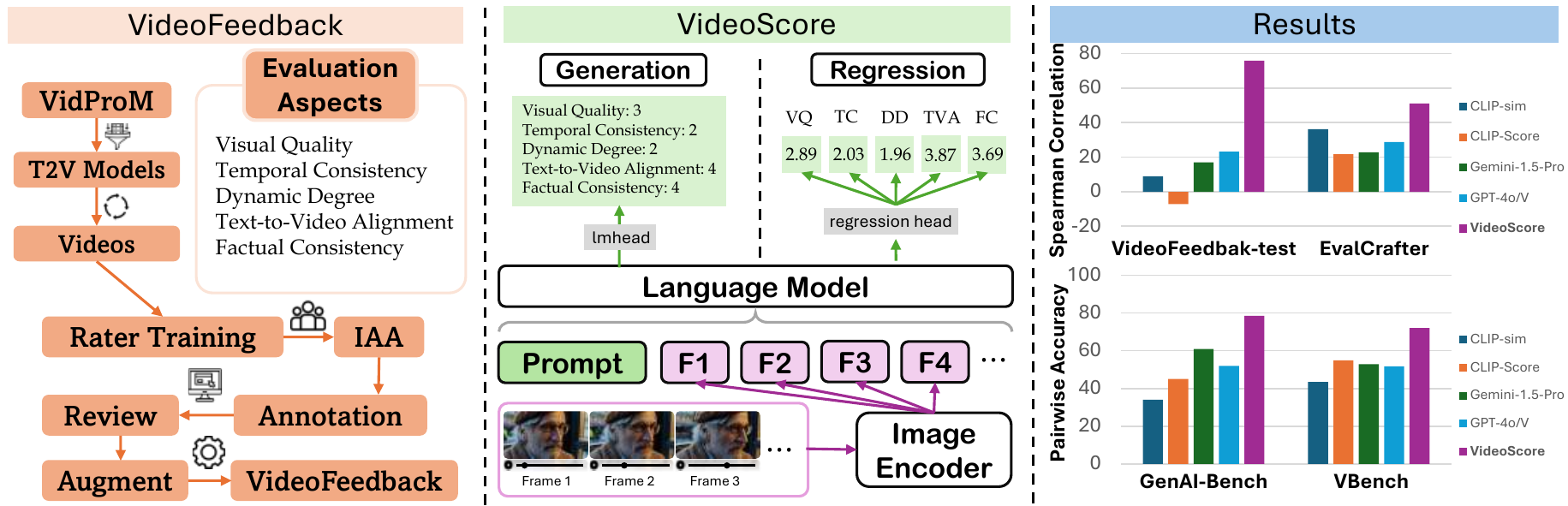}
        \captionof{figure}{Construction process of \data dataset and illustration of \score.}
        \vspace{1ex}
        \label{fig:teaser}
    \end{center}
}]

\begin{abstract}
The recent years have witnessed great advances in video generation. However, the development of automatic video metrics is lagging significantly behind. None of the existing metrics are able to provide reliable scores over generated videos. The main barrier is the lack of large-scale human-annotated datasets. In this paper, we release \data, the first large-scale dataset containing human-provided multi-aspect score over 37.6K synthesized videos from 11 existing video generative models. 
We train \score (initialized from Mantis) based on \data to enable automatic video quality assessment. 
Experiments show that the Spearman correlation between \score and humans can reach 77.1 on \data-test, beating the prior best metrics by about 50 points. Further results on other held-out EvalCrafter, GenAI-Bench, and VBench show that \score has consistently much higher correlation with human judges than other metrics. Due to these results, we believe \score can serve as a great proxy for human raters to (1) rate different video models to track progress (2) simulate fine-grained human feedback in Reinforcement Learning with Human Feedback (RLHF) to improve current video generation models. 
\end{abstract}
\vspace{1em}

\section{Introduction}
\label{sec:intro}

Powerful text-to-video (T2V) generative models have been exponentially emerging these days. In 2023 and 2024, we have witnessed an array of T2V models like Sora~\cite{OpenAI2023sora}, Runway Gen-2~\cite{esser2023structure}, Lumiere~\cite{bar2024lumiere}, Pika\footnote{https://pika.art/home}, Luma-AI\footnote{https://lumalabs.ai/dream-machine}, Kling\footnote{https://kling.kuaishou.com/}, Emu-video~\cite{girdhar2023emu}, StableVideoDiffusion~\cite{blattmann2023stable}. These models have shown their potential to generate longer-duration, higher-quality, and more natural videos. 
Despite the great progress video generative models have made, they still suffer from artifacts like unnaturalness, inconsistency and hallucination, which calls for reliable fine-grained metrics for evaluation and a robust reward model for reinforcement learning (RLHF).

The recent literature has adopted a wide range of metrics to evaluate videos. However, these metrics suffer from the following issues: (1) some of them are computed over distributions and cannot be adopted to evaluate a single model output. Examples include FVD~\cite{unterthiner2019fvd} and IS~\cite{salimans2016improved}. (2) most metrics can only be used to evaluate visual quality or text alignment, while failing on other aspects like motion smoothness, factual consistency, etc. Examples of such metrics include CLIP~\cite{radford2021learning}, DINO~\cite{caron2021emerging}, BRISQUE~\cite{mittal2012no}. (2) some metrics focus only on a single mean opinion score (MOS), failing to provide fine-grained subscores across different multiple aspects. Examples include T2VQA~\cite{kou2024subjective}, FastVQA~\cite{wu2022fast}, and DOVER~\cite{wu2023exploring}. (3) Several works~\cite{ku2023viescore,Bansal2024VideoPhyEP} propose to prompt multi-modal large-language-models (MLLM) like GPT-4o~\cite{achiam2023gpt} or Gemini-1.5~\cite{reid2024gemini} to produce multi-aspect quality assessment for given videos. However, our experiments show that they also have low correlation with humans. These feature-based metrics or MLLM prompting methods either fail to provide reliable evaluation or cannot simulate the human feedback from real world well, which have lagged behind and restricted us from training better video generative models.

Since obtaining large-scale human feedback is highly costly, we can try to approximate human-provided scores with model-based metrics. To this end, our work can be divided into two parts: (1) curating \data, the first large-scale dataset containing human-provided scores for 37.6K videos, (2) training \score on \data, which is an automatic video metric to simulate human feedback.

In preparation of \data, we solicit prompts from VidProM~\cite{Wang2024VidProMAM} and use 11 popular text-to-video models, including Pika, Lavie~\cite{wang2023lavie}, SVD~\cite{blattmann2023stable}, etc, to generate videos of various quality based on these prompts. We define five key aspects for evaluation as shown in \autoref{tab:aspects_definition}, and get our videos from \data annotated for each aspect from 1 (bad) to 4 (perfect).

To build the video evaluator, we select Mantis-Idefics2-8B~\cite{jiang2024mantis} as our main backbone model due to its superior ability to handle multi-image and video content, accommodating up to 128 video frames and supporting native resolution. After fine-tuning Mantis on \data-train, we get our video evaluator, \score.
Experiments show that we achieve a Spearman correlation of $77.1$ on \data-test and $59.5$ on EvalCrafter~\cite{liu2023evalcrafter} for the text-to-video alignment aspect, surpassing the best baseline by $54.1$ and $4.4$ respectively. The pairwise comparison accuracy gets $78.5$ on GenAI-Bench~\cite{jiang2024genai} video preference part, and $72.1$ in average on 5 aspects of VBench~\cite{Huang2023VBenchCB}, surpassing the previous best baseline by $11.4$ and $9.6$ respectively.
Additional ablation studies with different backbone models confirmed that the Mantis-based metric provides a gain of $12.1$ compared to using the Idefics2-based metric. Due to the significant improvement, we believe that \score can serve as the reliable metrics for future video generative models.

\section{Related Work}
\subsection{Text-to-Video Generative Models}
Recent progress in diffusion models~\cite{ho2020denoising, rombach2022high} has significantly pushed forward the development of Text-to-Video (T2V) generation. Given a text prompt, the T2V generative model can synthesize new video sequences that didn't previously exist~\cite{wang2023lavie, OpenAI2023sora, chen2023videocrafter1, chen2024videocrafter2, streamingt2v, bar2024lumiere}. Early diffusion-based video models generally build upon Text-to-Image (T2I) models, adding a temporal module to extend itself into the video domain~\cite{wang2023lavie, chen2023seine}. Recent T2V generation models are directly trained on videos from scratch. Among these, models based on Latent Diffusion Models (LDMs) have gained particular attention for their effectiveness and efficiency~\cite{zhou2022magicvideo, an2023latent, blattmann2023align}. While the other works used the pixel-based Diffusion Transformers (DiT) also achieve quality results~\cite{gupta2023photorealistic, chen2023gentron, OpenAI2023sora}.

\subsection{Video Quality Assessment}
As the current progress of Text-to-Video generative models leaves it uncertain how close we are to reaching the objective, researchers have worked on evaluation methods to benchmark the generative models. Common methods involve the use of FVD~\cite{unterthiner2018towards} and CLIP~\cite{radford2021learning_CLIP} to evaluate the quality of frames and the text-frames alignment respectively. However, other aspects like subject consistency, temporal consistency, factual consistency cannot be captured by these metrics. Recent works like VBench~\cite{Huang2023VBenchCB} proposes to use different DINO~\cite{caron2021emerging}, optical flow~\cite{horn1981determining} to reflect these aspects. However, the correlation with human judgment is relatively low. For example, most models have subject/background consistency scores over 97\% in VBench, which is a massive overestimation of the current T2V models' true capability.  Another work EvalCrafter~\cite{liu2023evalcrafter} instead resorts to human raters to perform comprehensive evaluation. 

A recent work VideoPhy~\cite{Bansal2024VideoPhyEP} follows VIEScore~\cite{ku2023viescore} prompt large multi-modal models like Gemini~\cite{reid2024gemini} and GPT-4o~\cite{achiam2023gpt} to provide quality assessment. However, our later study shows that these multimodal language models also achieve very low agreement with human raters. A concurrent work T2VQA~\cite{Kou2024SubjectiveAlignedDA} also proposes to train a quality assessment model on human-annotated video ratings. However, there are a few distinctions. Firstly, our dataset contains ratings for multiple aspects. Secondly, our dataset is 4x larger than the T2VQA dataset. Thirdly, our metric is built on pre-trained video-language foundation models to maximize its performance.

\subsection{RLHF in image/video generation}
In recent years, reinforcement learning from human feedback (RLHF) has emerged as a significant approach to enhancing the performance of image/video generative models. Numerous studies have focused on training reward models with large datasets of image-text pairs, such as HPSv2~\cite{wu2023human}, ImageReward~\cite{Xu2023ImageRewardLA}, RichHF-18K~\cite{Liang2023RichHF}, or video-text pairs like T2V-Score~\cite{Wu2024TowardsAB}. Moreover, some research efforts have combined existing reward models to simulate human feedback, such as T2V-Turbo~\cite{Li2024T2VTurboBT}, while some recent works (\citet{Wang2024InterpretablePV}, \citet{Wang2024ArithmeticCO}) proposed multi-objective reward model with regression head to provide human preferences. Utilizing these reward models or feedback simulators, diverse methods have been proposed to align the output of visual generative models with human preferences, including RL-based methods (\citet{fan2024reinforcement}, \citet{Zhang2024LargescaleRL}) and reward fine-tuning methods (\citet{Clark2023DirectlyFD}, \citet{Li2024RewardGL}, \citet{Yuan2023InstructVideoIV}). Additionally, some works adopt data distillation to fine-tune diffusion models on high-quality data, while others like Diffusion-DPO~\cite{Wallace2023DiffusionMA}, extend Direct Preference Optimization (DPO) to train diffusion models based on preference data. Our \score aims to approximate human feedback, which is expected to be beneficial in enhancing video generative models with RLHF methods like PPO or DPO.

\section{\data}
\label{sec:dataset}
\begin{table*}[!ht]
  \small
  \centering
  \vspace{1em}
  \begin{tabular}{lcccccc}
    \toprule
    Base Model or Video Type  & Video Source  & Total Size & Resolution & Duration & FPS & Score \\
    \midrule
    \multicolumn{7}{c}{Human Annotated Videos} \\
    \midrule
    Pika                      & VidProM & 4.6k & (768, 480) & 3.0s & 8 & [1-4] \\
    Text2Video-Zero~\cite{text2video-zero}        & VidProM & 4.6k & (512,512) & 2.0s & 8 & [1-4]  \\
    VideoCrafter2~\cite{chen2024videocrafter2} & VidProM & 4.9k& (512, 320)& 2.0s & 8 & [1-4] \\
    ModelScope~\cite{Wang2023ModelScopeTT} & VidProM & 4.5k& (256, 256) & 2.0s & 8 & [1-4] \\
    LaVie-base~\cite{wang2023lavie} & Generated & 3.2k& (512, 320)& 2.0s & 8 & [1-4] \\
    AnimateDiff~\cite{guo2023animatediff} & Generated & 1.4k& (512, 512)& 2.0s & 8 & [1-4]  \\
    LVDM~\cite{he2022lvdm} & Generated & 3.1k& (256, 256)& 2.0s & 8  & [1-4] \\
    Hotshot-XL~\cite{MullanHotshot} & Generated & 3.2k& (512, 512)& 1.0s & 8 & [1-4]  \\
    ZeroScope-576w~\cite{zeroscope} & Generated & 2.2k& (256, 256)& 2.0s & 8 & [1-4]  \\
    Fast-SVD~\cite{blattmann2023stable} & Generated & 1.0k& (1024, 576)& 3.0s & 8 & [1-4] \\
    SoRA-Clip~\cite{OpenAI2023sora} & Collected & 0.9k& various & 2.0/3.0s & 8  & [1-4]\\
    \midrule
    \multicolumn{7}{c}{Augmented Videos} \\
    \midrule
    DiDeMo~\cite{hendricks2017localizing} & Real & 2.0k& various & 2.0/3.0s & 8 & 4\\
    Panda70M~\cite{chen2024panda70m}      & Real & 2.0k& various & 2.0/3.0s & 8 & 4 \\
    \bottomrule
  \end{tabular}
\caption{Statistics of our curated \data for training video-generation evaluator. It consists of 33.6K human-scored videos across multiple aspects, with 4k real-world videos collected from DiDeMo~\cite{hendricks2017localizing} and Panda70M~\cite{chen2024panda70m} as the supplementary data. Ultimately, we get 37.6K high-quality rated videos as the final \data.}
\label{tab:dataset_statistics}
\end{table*}

This section introduces the construction process of our dataset, \data. We start by explaining how we gathered and filtered diverse text prompts for video generation, followed by the video-generation processes using 11 selected text-to-video models. Next, we outline the annotation pipeline that guides raters to score videos across multiple aspects defined in ~\autoref{tab:aspects_definition}. We also include supplementary data to enhance robustness. Finally, we summarize the dataset statistics in ~\autoref{tab:dataset_statistics}, with 760 examples as the test set.

\subsection{Data preparation}
\label{subsec:data_preparition}
\paragraph{Prompt Sources} 
We utilize VidProM~\cite{Wang2024VidProMAM}, a dataset containing extensive text-to-video pairs from different models. VidProM's video-generation prompts are diverse and semantically rich, derived from real-world user inputs. To create a manageable subset from the 1.04 million unique prompts, we apply two filters: a length filter and an NSFW filter. The length filter eliminates prompts with fewer than 5 words or more than 100 words. The NSFW filter removes prompts with a high probability of containing inappropriate content. After filtering, we perform random down-sampling to obtain a set of 44.5K prompts, 31.6K of them are used in video generation and some videos may have the same text prompt.

\paragraph{Video Generation} 
We select 11 text-to-video (T2V) generative models (shown in ~\autoref{tab:dataset_statistics}) with various capabilities so that the quality of the generated video ranges from high to low in a balanced way. Some videos are pre-generated in the VidProM dataset, including Pika, Text2Video-Zero~\cite{text2video-zero}, VideoCrafter2~\cite{chen2024videocrafter2}, and ModelScope~\cite{Wang2023ModelScopeTT}, whereas the others are generated by ourselves or collected from the Internet (i.e. SoRA). To eliminate differences between models in subsequent annotation stage, we normalize the videos into a unified format. First, we standardized the frame rate to 8 fps to address discrepancies in temporal consistency between high and low fps videos. Specifically, for high frame rate model Pika and AnimateDiffusion~\cite{guo2023animatediff} we use frame down sampling, while for low frame rate model like Text2Video-Zero, we employed frame interpolation~\cite{huang2022realtime} on it. Details are shown in ~\autoref{subsec:video_format_norm}. Additionally, we cropped Pika videos to remove the watermark, making them indistinguishable from other models. Ultimately, we obtained 33.6K videos from 11 T2V models, along with their generation prompts.

\subsection{Annotation Pipeline}
\label{subsec:annotation_pipeline}
\begin{table*}[]

\small
\centering
\begin{tabular}{ll}
\toprule
Aspect                   & Definition                                                                        \\
\midrule
Visual Quality (VQ)   & the quality of the video in terms of clearness, resolution, brightness, and color \\
Temporal Consistency (TC)     & the consistency of objects or humans in video                                     \\
Dynamic Degree (DD)          & the degree of dynamic changes                                                     \\
Text-to-Video Alignment (TVA) & the alignment between the text prompt and the video content                       \\
Factual Consistency (FC)     & the consistency of the video content with common-sense and factual knowledge  \\
\bottomrule
\end{tabular}
\caption{The five evaluation aspects of \data and their definitions.}
\label{tab:aspects_definition}
\end{table*} 

\begin{table}[!ht]

\small
\centering
\begin{tabular}{l|ccccc}
\toprule
IAA metric  & VQ    & TC    & DD    & TVA   & FC     \\
\midrule
\multicolumn{6}{c}{Trial 1 (\#=30)}                  \\
\midrule
Match Ratio & 0.733 & 0.706 & 0.722 & 0.678 & 0.633  \\
Kappa       & 0.369 & 0.414 & 0.413 & 0.490 & 0.265  \\
Alpha       & 0.481 & 0.453 & 0.498 & 0.540 & 0.365  \\
\midrule
\multicolumn{6}{c}{Trial 2 (\#=100)}                 \\
\midrule
Match Ratio & 0.787 & 0.699 & 0.913 & 0.570 & 0.727  \\
Kappa       & 0.088 & 0.562 & 0.565 & 0.125 & -0.089 \\
Alpha       & 0.078 & 0.579 & 0.620 & 0.205 & -0.106 \\
\bottomrule
\end{tabular}
\caption{Inter-Annotator Agreement (IAA) analysis results considering Matching Ratio, Fleiss' $\kappa$, and Krippendorff's $\alpha$ on the two trial annotations.}
\label{tab:IAA}
\end{table}

\paragraph{Evaluation Dimensions}
As discussed in ~\autoref{sec:intro}, fine-grained and multi-aspect rating of videos is crucial for enhancing both the reliability and explainability of the video evaluator. Inspired by VBench~\cite{Huang2023VBenchCB} and EvalCrafter~\cite{liu2023evalcrafter}, and FETV~\cite{liu2023fetv}, we propose five key dimensions for text-to-video evaluation, detailed in ~\autoref{tab:aspects_definition}. These dimensions encompass both low-level vision aspects, such as Visual Quality, which evaluates basic visual impressions, and higher-level aspects, like Text-to-Video Alignment and Factual Consistency, which require a deep understanding of world knowledge, is a capability previous metrics do not have. Besides definition, a checklist for error points for each dimension is also provided to assist the rater in contributing more accurate and consistent rating. Detailed are provided in~\autoref{tab:aspect_error_points}. 

\paragraph{Annotation}
We hired 20 expert raters, with each rater performing rating for 1K-2K videos. Our raters are mostly college graduate students.  For each aspect, there are three available ratings, 1 (Bad), 2 (Average), and 3 (Good), the score 4 (Perfect) is post-annotated, as described in the ~\autoref{subsec:data_augmentation}. To ensure the consistency and quality of the annotations, we conducted a system training for each rater. Initially, we conducted a pilot training session with examples of multi-aspect ratings for various videos. Following this, multiple rounds of small-scale annotation were conducted to compute the inter-annotator agreement (IAA) across five aspects, as shown in ~\autoref{tab:IAA}. The results indicate a high score-matching ratio for all aspects, along with Fleiss' $\kappa$~\cite{fleiss1973equivalence} and Krippendorff's $\alpha$~\cite{Krippendorff2011ComputingKA} metrics, with values around 0.4 or 0.5, suggesting sufficient agreement to proceed with large-scale annotation. The annotation process takes roughly 4 weeks to finish. 

\paragraph{Review}
We conduct random checks on different raters during the annotating phase to ensure the alignment. Once we find the exceeded unqualified ratio in certain rater, we promptly communicate with the respective rater and review the annotations for that segment of the video. This helps calibrate the annotation provided by that rater during the relevant period. For example, we found several raters are too lenient and tend to give high scores to unqualified videos. We then step in to make sure they are aligned with our understanding of evaluation dimensions.
With periodical random inspection on annotating, we completed the large-scale annotation of 33.6K videos and moved to the data augmentation stage.

\begin{figure}
    \centering
    \vspace{-1em}
    \includegraphics[scale=0.24]{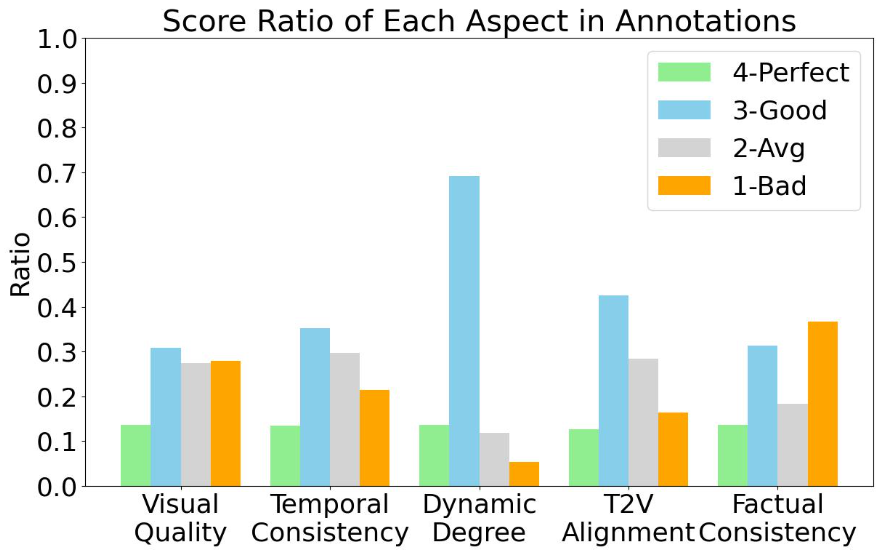}
    \caption{The rating distribution on all the videos.}
    \label{fig:rating_dist}
    \vspace{-1em}
\end{figure}

\subsection{Dataset Augmentation}
\label{subsec:data_augmentation}
To enhance the robustness of our dataset, we incorporated post-augmentation into the dataset. Firstly, expert raters will review the excellent videos (all aspects are scored 3) again to select perfect ones and raise their scoring to 4 (Perfect) in certain aspects, particularly among the SoRA and FastSVD~\cite{blattmann2023stable} videos.

Additionally, we gather 4k real-world videos from the DiDeMo~\cite{hendricks2017localizing} and Panda70M~\cite{chen2024panda70m} with each video accompanied by a text description. We select and cut clips from the ones less than 5 seconds to ensure a strong match between video and its text. 
We apply similar normalization in ~\autoref{subsec:data_preparition} and also use SSIM and
 MSE between interval sampled frames to filter out the possible static videos, ensuring the quality in Dynamic Degree. Finally the 4K real videos are scored 4 (perfect) in all aspects. 

We plot the rating distributions across each dimension in ~\autoref{fig:rating_dist}. which is balanced except for Dynamic Degree. We inspected in detail via case study and turned out this distribution is expected. Eventually, we get the final 37.6K examples as the training split of \data, and reserve 760 validation examples as test set. 

\section{Experiments}
\label{sec:experiments}

In this section, we describe our experiment setup, including baseline methods for video evaluation, and evaluation benchmarks for video evaluation. We also discuss the training details of \score, and analyze our experiment results. 

\begin{table*}[!ht]
\small
\centering
\vspace{-1em}
\begin{tabular}{l|ccccc|c}
\toprule
Method & Visual Quality& Temporal & Dynamic Degree& Text Alignment & Factual & Avgerage \\
\midrule
Random & -3.1 & 0.5 & 0.4 & 1.1 & 2.9 & 0.4 \\
\midrule
\multicolumn{7}{c}{\textbf{Feature-basd automatic metrics}} \\
\midrule
PIQE & -17.7 & -14.5 & 1.2 & -3.4 & -16.0 & -10.1 \\
BRISQUE & -32.4 & -26.4 & -4.9 & -8.6 & -29.1 & -20.3 \\
CLIP-sim & 21.7 & 29.1 & -34.4 & 2.0 & 26.1 & 8.9 \\
DINO-sim & 19.4 & 29.6 & -37.9 & 2.2 & 24.0 & 7.5 \\
SSIM-sim & 33.0 & \underline{30.6} & -31.3 & 4.7 & 30.2 & 13.4 \\
MSE-dyn & -20.3 & -24.7 & \underline{38.0} & 3.3 & -23.9 & -5.5 \\
SSIM-dyn & -31.4 & -29.1 & 31.5 & -5.3 & -30.0 & -12.9 \\
CLIP-Score & -10.9 & -10.0 & -14.7 & -0.3 & -0.3 & -7.2 \\
X-CLIP-Score & -3.2 & -2.7 & -7.3 & 5.9 & -2.0 & -1.9 \\
\midrule
\multicolumn{7}{c}{\textbf{MLLM Propmting}} \\
\midrule
LLaVA-1.5-7B & 9.4 & 8.0 & -2.2 & 11.4 & 15.8 & 8.5 \\
LLaVA-1.6-7B & -8.0 & -4.1 & -5.7 & 1.4 & 0.8 & -3.1 \\
Idefics2 & 4.2 & 4.5 & 8.9 & 10.3 & 4.6 & 6.5 \\
Gemini-1.5-Flash & 24.1 & 5.0 & 20.9 & 21.3 & \underline{32.9} & 20.8 \\
Gemini-1.5-Pro & \underline{35.2} & -17.2 & 18.2 & \underline{26.7} & 21.6 & 16.9 \\
GPT-4o & 13.6 & 17.6 & 28.2 & 25.7 & 30.2 & \underline{23.0} \\
\midrule
\multicolumn{7}{c}{\textbf{Ours}} \\
\midrule
\score (gen) & \textbf{86.2} & 80.3 & \textbf{77.6} & 59.4 & 82.1 & \textbf{77.1} \\
\score (reg) & 84.7 & \textbf{81.5} & 68.4 & \textbf{59.5} & \textbf{84.6} & 75.7 \\
\rowcolor{LightCyan} $\Delta$ over Best Baseline & +51.0 & +50.9 & +39.6 & +32.8 & +51.7 & +54.1 \\
\bottomrule
\end{tabular}
\caption{Correlation (Spearman's $\rho$) between model answer and human reference on \data-test. }
\label{tab:test_videoeval}
\vspace{-3ex}
\end{table*}

\subsection{Baselines}
\label{baselines}
To compare with our evaluator model, we selected two categories of video quality metrics. The first category relies on statistical or neural features for evaluation. These metrics typically assess a single video dimension such as temporal consistency, and then yield a numerical value. The second category employs advanced MLLMs to evaluate videos across multiple dimensions. Extensive literature demonstrates that MLLMs not only excel in generating content on user instructions but also outperform traditional metrics in evaluating AI-generated content (AIGC). All baselines are listed in ~\autoref{tab:test_videoeval}.

\paragraph{Feature-Based Metrics} We list all the experimented metrics as follows:
\begin{enumerate}[itemsep=0pt]
  \item Visual Quality.
  We use two no-reference image quality metrics PIQE~\cite{piqe} and BRISQUE~\cite{brisque}. We apply them on all frames of video and take the average score across frames. 
  \item Temporal Consistency.
  In this dimension, CLIP-sim~\cite{radford2021learning} and DINO-sim~\cite{caron2021emerging}  are computed as cosine similarities of between adjacent frames features, following VBench~\cite{Huang2023VBenchCB}. Additionally, we calculate SSIM between adjacent frames, denoted as SSIM-sim.
  \item Dynamic Degree.
  We uniformly sample four frames from the target video and calculate the average MSE (Mean Square Error) and SSIM~\cite{ssim_1284395} between adjacent frames in the sample as final score.
  \item Text-to-Video Alignment.
  We include CLIP-Score~\cite{radford2021learning} and X-CLIP-Score~\cite{ma2022x} as metrics in this dimension. CLIP-Score calculates cosine similarity between the feature of each frame and the text prompt and then averages across all frames, while X-CLIP-Score utilizes the feature of video instead of frames.
  \item Factual Consistency.
  It is challenging to find a feature-based metric to determine whether the visual content aligns with common sense. Therefore, we rely on the second category of metrics for this dimension.
\end{enumerate}

We discretized the continuous outputs of these metrics to align with our labeling scores [1, 2, 3, 4]. For instance, for CLIP-sim, values are converted to: '4' if raw output in [0.97, 1], '3' if in  [0.9, 0.97), '2' if in [0.8, 0.9) and '1' otherwise. See ~\autoref{tab:dis_rules} for more details.

\paragraph{MLLM Prompting Based Metrics}
To understand how existing MLLMs perform on the multi-aspect video evaluation task, we designed a prompting template in ~\autoref{tab:mllm_prompt_template_gen} to let them output scores ranging from 1 (Bad) to 4 (Perfect) for each aspect. However, some models, including Idefics2~\cite{Laurenon2024WhatMW}, Fuyu~\cite{fuyu8b}, Kosmos-2~\cite{Peng2023Kosmos2GM}, and CogVLM~\cite{Wang2023CogVLMVE} and OpenFlamingo~\cite{awadalla2023openflamingo}, fail to give reasonable outputs. We thus exclude them from the tables. MLLMs that follow the output format like LLaVA-1.5~\cite{Liu2023VisualIT}, LLaVA-1.6~\cite{liu2024llavanext}, Idefics1~\cite{laurencon2023obelics}, Google's Gemini 1.5~\cite{reid2024gemini}, and OpenAI's GPT-4o~\cite{openai2024gpt4o}. 

\subsection{Evaluation Benchmarks}
We have included the following benchmarks to evaluate the ability of \score and the above-mentioned baselines on evaluating model generated videos to see their correlation with human raters. 

\paragraph{\data-test}
As mentioned in ~\autoref{sec:dataset}, we split 760 video entries from \data dataset, which contains 680 annotated videos and 80 augmented videos. We take label prediction accuracy and Spearman's $\rho$ in each dimension as evaluation indicators. For a specific aspect in the this set (e.g. Visual Quality), we use the predicted score from the same aspect to measure the performance for baselines and our models.

\paragraph{GenAI-Bench}
GenAI-Bench~\cite{jiang2024genai} is a benchmark designed to evaluate MLLM's ability on preference comparison for tasks including text-to-video generation and others. The preference data is taken from GenAI-Arena from user voting. We select the video preference data in our experiments.
This involves the MLLM judging which of the two provided videos is generally better, measured by pairwise accuracy. We use the averaged scores of the five aspects for MLLM prompting baselines and our models to give the preference. We compute the correlation between model-assigned preference vs. human preference as our indicator. 

\begin{table*}[!t]
\small
\centering
\vspace{-1em}
\begin{tabular}{l|c|ccccc}
\toprule
Benchmark $\rightarrow$ & GenAI-Bench & \multicolumn{5}{c}{VBench} \\
\midrule
Model $\downarrow$ Sub-Aspect $\rightarrow$ & \begin{tabular}[c]{@{}c@{}}Video \\ Preference\end{tabular} & \begin{tabular}[c]{@{}c@{}}Technical\\ Quality\end{tabular} & \begin{tabular}[c]{@{}c@{}}Subject\\ Consistency\end{tabular} & \begin{tabular}[c]{@{}c@{}}Dyanmic\\ Degree\end{tabular} & \begin{tabular}[c]{@{}c@{}}Motion\\ Smoothness\end{tabular} & \begin{tabular}[c]{@{}c@{}}Overall\\ Consistency\end{tabular} \\
\midrule
Random & 37.7 & 44.5 & 42.0 & 37.3 & 40.5 & 44.8 \\
\midrule
\multicolumn{7}{c}{\textbf{Feature-based Automatic Metrics}} \\
\midrule
PIQE & 34.5 & \underline{60.8} & 44.3 & 71.0 & 45.3 & 53.8 \\
BRISQUE & 38.5 & 56.7 & 41.2 & 75.5 & 41.2 & 54.2 \\
CLIP-sim & 34.1 & 47.8 & 46.0 & 34.8 & 44.7 & 44.2 \\
DINO-sim & 31.4 & 49.5 & \underline{51.2} & 24.7 & \underline{55.5} & 41.7 \\
SSIM-sim & 28.4 & 30.7 & 46.2 & 24.5 & 54.2 & 27.2 \\
MSE-dyn & 34.2 & 32.8 & 31.7 & 81.7 & 31.2 & 39.2 \\
SSIM-dyn & 38.5 & 37.5 & 36.3 & \underline{84.2} & 34.7 & 44.5 \\
CLIP-Score & 45.0 & 57.8 & 46.3 & 71.3 & 47.0 & 52.2 \\
X-CLIP-Score & 41.4 & 44.0 & 38.0 & 51.0 & 28.7 & 39.0 \\
\midrule
\multicolumn{7}{c}{\textbf{MLLM Prompting}} \\
\midrule
LLaVA-1.5-7B & 49.9 & 42.7 & 42.3 & 63.8 & 41.3 & 8.8 \\
LLaVA-1.6-7B & 44.5 & 38.7 & 26.8 & 56.5 & 28.5 & 43.2 \\
Idefics1 & 34.6 & 20.7 & 22.7 & 54.0 & 27.3 & 33.7 \\
Gemini-1.5-Flash & 67.1 & 52.3 & 49.2 & 64.5 & 45.5 & 49.9 \\
Gemini-1.5-Pro & \underline{60.9} & 56.7 & 43.3 & 65.2 & 43.0 & 56.3 \\
GPT-4o & 52.0 & 59.3 & 49.3 & 46.8 & 42.0 & \underline{60.8} \\
\midrule
\multicolumn{7}{c}{\textbf{Ours}} \\
\midrule
\score (gen) & 59.0 & 64.2 & 57.7 & 55.5 & 54.3 & 61.5 \\
\score (reg) & \textbf{78.5} & \textbf{78.2} & \textbf{71.5} & \textbf{68.0} & \textbf{74.0} & \textbf{69.0} \\
\rowcolor{LightCyan} $\Delta$ over Best Baseline & +11.4 & +17.4 & +20.3 & -16.2 & +18.5 & +8.2 \\
\bottomrule
\end{tabular}
\caption{
Pairwise preference accuracy on GenAI-Bench~\cite{jiang2024genai} and VBench~\cite{Huang2023VBenchCB}. For MLLM prompting and our method, we averaged the five aspect scores defined in Table~\ref{tab:aspects_definition} as the score for each video in the comparison, where the higher one deemed the winner. 
}
\label{tab:test_genaibench_vbench}
\vspace{-1em}
\end{table*}

\paragraph{VBench}
VBench~\cite{Huang2023VBenchCB} is a comprehensive multi-aspect benchmark suite for video generative models, where they use a bunch of existing auto-metrics in each aspect. VBench have released a set of human preference annotations on all the aspects, comprising videos by 4 models, including ModelScope~\cite{Wang2023ModelScopeTT}, CogVideo~\cite{hong2022cogvideo}, VideoCrafter1~\cite{chen2023videocrafter1}, and LaVie~\cite{wang2023lavie}. We select the subset from 5 aspects of VBench, like technical quality, subject consistency, and so on, to compute the preference comparison accuracy. For each aspect, we subsample 100 unique prompts in the testing. We use the averaged scores of the five aspects for MLLM prompting baselines and our models to predict the preference.

\paragraph{EvalCrafter}
EvalCrafter~\cite{liu2023evalcrafter} is a text-to-video benchmark across four dimensions: Video Quality, Temporal Consistency, Text-to-Video Alignment, and Motion Quality. We focused on the first three ones and gathered 2,541 videos by five models: Pika, Gen2, Floor33~\cite{floor33}, ModelScope, and ZeroScope~\cite{zeroscope}. In EvalCrafter, human annotators rated each video on a scale of 1-5, with each scored by three raters. We calculated the average score across raters and normalized it to [0, 1]. After inference on benchmark videos, we excluded "Dynamic Degree" and "Factual Consistency" to match EvalCrafter's dimensions. Finally, we used Spearman's $\rho$ in each dimension as an indicator.

\begin{table}[!hb]
\small
\centering
\vspace{-1ex}
\begin{tabular}{l|ccc}
\toprule
Method & Visual & Temporal & Text Align \\
\midrule
Random & -2.0 & 1.4 & -0.9 \\
EvalCrafter & \underline{55.4} & \underline{56.7} & 32.3 \\
\midrule
\multicolumn{4}{c}{Feature-based Automatic Metrics} \\
\midrule
PIQE & 0.5 & -3.3 & -0.9 \\
BRISQUE & 6.4 & -1.3 & 6.7 \\
CLIP-sim & 36.0 & 53.5 & 19.2 \\
DINO-sim & 30.6 & 50.3 & 15.3 \\
SSIM-im & 32.4 & 36.9 & 11.4 \\
MSE-dyn & -15.4 & -27.5 & -8.1 \\
SSIM-dyn & -32.6 & -33.9 & -12.6 \\
CLIP-Score & 18.7 & 11.5 & 35.0 \\
X-CLIP-Score & 12.2 & 3.1 & 24.5 \\
\midrule
\multicolumn{4}{c}{MLLM Prompting} \\
\midrule
LLaVA-1.5-7B & 13.4 & 15.6 & 2.6 \\
LLaVA-1.6-7B & 12.2 & 8.5 & 18.9 \\
Idefics1 & 1.5 & -1.5 & 0.8 \\
Gemini-1.5-Flash & 34.9 & -27.8 & 44.8 \\
Gemini-1.5-Pro & 37.8 & -24.1 & \underline{55.1} \\
GPT-4o & 32.9 & 12.5 & 40.7 \\
\midrule
\multicolumn{4}{c}{Ours} \\
\midrule
\score (gen) & 20.8 & 51.3 & 10.7 \\
\score (reg) & \textbf{42.4} & \textbf{51.3} & \textbf{59.5} \\
\rowcolor{LightCyan} $\Delta$ over Best Baseline & -13.1 & -5.4 & 4.4 \\
\bottomrule
\end{tabular}
\caption{Spearman's Correlation ($\rho$) of \score on EvalCrafter~\cite{liu2023evalcrafter}. }
\label{tab:test_evalcrafter}
\end{table}

\begin{table*}[!t]
\small
\centering
\vspace{-1em}
\begin{tabular}{lc|cccc|c}
\toprule
Base Model & Scoring Type & \data$^*$ & EvalCrafter$^*$ & GenAI-Bench & VBench$^*$ & Average \\
\midrule
VideoLLaVA-7B & Generation & 71.9 & 9.8 & 42.6 & 46.5 & 42.7 \\
Idefics2-8B & Generation & 73.9 & 11.3 & 50.7 & 53.9 & 47.5 \\
Mantis-Idefics2-8B & Generation & {\ul 77.1} & {\ul 27.6} & 59.0 & 58.7 & 55.6 \\
Idefics2-8B & Regression & 73.9 & 17.4 & {\ul 74.5} & {\ul 64.4} & 57.5 \\
Mantis-Idefics2-8B & Regression & \textbf{75.7} & \textbf{51.1} & \textbf{78.5} & \textbf{73.0} & \textbf{69.6} \\
\bottomrule
\end{tabular}
\caption{Ablation study on the base model and scoring function for \score. "$^*$" means that we take the average of Spearman correlation or pairwise accuracy across the multiple aspects of the benchmark. The highest numbers are bold for each benchmark, and the second are underlined.}
\label{tab:model_ablation}
\vspace{-0em}
\end{table*}


\subsection{Training Details}
For \score, We use two scoring methods: generative scoring and regression scoring. Generative scoring involves training the model to output fixed text forms, from which aspect scores are extracted using regular expressions. These scores are integers corresponding to human annotation scores. In contrast, regression scoring replaces the language model head with a linear layer that outputs 5 logits representing scores for each aspect. Regression scoring is trained using MSE loss.

We select Mantis-Idefics2-8B~\cite{jiang2024mantis} as the base model, which can accommodate 128 video frames at most. The learning rate is set to 1e-5. Each model is trained for 1 epoch on 8 A100 (80G) GPUs, finishing in 6 hours.

\subsection{Evaluation Results}
We report the Spearman correlation results on the \data-test and EvalCrafter in ~\autoref{tab:test_videoeval} and ~\autoref{tab:test_evalcrafter}, respectively. For the preference comparison on videos, we report the pairwise accuracy on the GenAI-Bench and VBench in ~\autoref{tab:test_genaibench_vbench}. 

\paragraph{\score achieves the SoTA performance}
On the \data-test, \score gets an average of $54.1$ improvements on all the five aspects compared to the baseline GPT-4o. What's more, on the EvalCrafter benchmark, \score (reg) has $4.4$ improvements on text-to-video alignment. For pairwise preference comparison, \score also gets $78.5$ accuracy on GenAI-Bench, surpassing the second-best Gemini-1.5-Flash by $11.4$ points. on the Vbench, our model archives the highest pairwise accuracy on 4 out of 5 aspects from VBench, with an average of $16.1$ improvements.

\paragraph{Feature-based Automatic Metrics are limited}
While some feature-based automatic metrics are good at a single aspect, they might fail to evaluate well on others. For example, on the \data-test, the correlation scores of SSIM-dyn and MSE-dyn achieve $31.5$ and $38.0$ for the dynamic degree aspect, but they both get a negative correlation for others. Besides, PIQE, BRISQUE, CLIP-Score, and X-CLIP-Score get nearly all negative correlations for all 5 aspects. This proves the image quality assessment metrics cannot be easily adapted to the video quality assessment task.

\subsection{Best-of-K Sampling with \score}
While best-of-K sampling has proven effective in boosting the performance of LLMs~\cite{alpaca_eval}, it's still unknown whether it also works for video generation tasks. To investigate this problem and also better show the effectiveness of \score, we conduct a comparison of different T2V models \textbf{with} and \textbf{without} employing best-of-k sampling with \score on EvalCrafter.

We set $k=5$ and generated videos using 700 prompts from EvalCrafter. For each prompt, the video with the highest average \score across five dimensions was selected. We then evaluated both the "best videos" and randomly chosen ones using EvalCrafter's metrics, averaging the results over 700 videos to obtain the model's final score. 
As shown in ~\autoref{tab:best-of-k}, compared to the random sample, most scores on the EvalCrafter benchmark have increased after the best-of-5 process.

\renewcommand{\arraystretch}{1.25} 
\begin{table*}[!t]
\small
\centering
\begin{tabular}{l|cc|cc|cc|cc|cc}
\toprule
\multirow{3}{*}{T2V Model}  &  \multicolumn{10}{c}{Dimensions of EvalCrafter}  \\ 
\cline{2-11}
 & \multicolumn{2}{c|}{Average} & \multicolumn{2}{c|}{Visual } & \multicolumn{2}{c|}{Temporal} & \multicolumn{2}{c|}{Motion} & \multicolumn{2}{c}{T2V Align} \\
       & random & best &  random & best & random & best & random & best & random & best\\ 
\midrule

AnimateDiff 
& \textbf{62.16} & 61.87 
& 60.79 & \textbf{60.89} 
& 60.69 & \textbf{60.94 }
& 54.35 & \textbf{54.83 }
& \textbf{72.79} & 70.83\\

HotShot-XL 
& 53.69 &  \textbf{61.22}
& 57.56 &  \textbf{60.39}
& 58.85 &  \textbf{60.94}
& 51.52 &  \textbf{54.05}
& 46.83 &  \textbf{69.52}\\

LaVie-base 
& 57.88 &  \textbf{59.90}
& 57.79 &  \textbf{58.52}
& 54.21 &  \textbf{57.70}
& 52.51 &  \textbf{53.53}
& 66.99 & \textbf{69.86}\\

VideoCrafter2 
& 58.31 &  \textbf{58.98}
& 58.44 &  \textbf{59.70}
& 59.18 &  \textbf{60.79}
& 54.39 &  \textbf{54.65}
& \textbf{61.23}  & 60.77\\

VideoCrafter1 
& 54.30 & \textbf{57.28}
& 52.40 &  \textbf{53.68}
& 56.48 &  \textbf{59.81}
& 54.01 &  \textbf{54.88}
& 54.32 &  \textbf{60.76}\\

ModelScope 
& 52.45 & \textbf{54.48 }
& 44.80 & \textbf{45.01 }
& 56.70 &  \textbf{60.34}
& 53.20 &  \textbf{54.42}
& 55.09 & \textbf{58.17}\\

ZeroScope-576w 
& 51.07 &  \textbf{54.09}
& 43.36 &  \textbf{43.82}
& 55.98 &  \textbf{58.74}
& 54.55 &  \textbf{54.68}
& 50.39 & \textbf{59.12}\\
LVDM 
& 45.80 &  \textbf{46.04}
& 44.45 &  \textbf{44.64}
& 40.44 &  \textbf{43.09}
& \textbf{53.68} &  53.26
& \textbf{44.61} & 43.16\\

\bottomrule
\end{tabular}
\caption{Performace of T2V models on EvalCrafter \textbf{with} and \textbf{without} best-of-5 sampling using \score. Most EvalCrafter scores have increased compared to the random sample, proving the effectiveness of \score}
\label{tab:best-of-k}
\end{table*}

\subsection{Ablation Study}
We conducted an ablation study on the base model selection and scoring types by training different variants on \data. Results of the ablation study are shown in~\autoref{tab:model_ablation}.

\paragraph{Base model ablation}
To investigate the effects of changing the base model, we choose VideoLLaVA-7B and Idefics2-8B as base models from recent popular VLMs (\citet{Lin2023VideoLLaVALU}, \citet{Laurenon2024WhatMW}, \citet{Zhang2023VideoLLaMAAI}, \citet{li2024llavanext-ablations}). Since \data-test, EvalCrafter, and VBench both have multiple aspects in the benchmarks, we take the average score across these aspects and report the general performance in ~\autoref{tab:model_ablation}. The results show that the Video-LLaVA-based version gets the worst performance on the four benchmarks, even if it is specifically designed for video understanding. The Idefics2-8B-based version has marginal improvements compared to the VideoLLaVA. After changing to Mantis-Idefics2-8B, the scores on the four benchmarks keep improving from $47.5$ to $55.6$ on average. When the scoring type is regression, the mantis-based version is still better than the Idefics2-based version by $12.1$ points. Therefore, we select the Mantis-based version as the final choice.

\paragraph{Regression scoring or generative scoring?}
The primary difference between regression scoring and generative scoring is that regression scoring can give more fine-grained scores instead of just the four labels. Results on EvalCrafter, GenAI-Bench, and VBench all indicate that using regression scoring can consistently improve the Spearman correlation or the pairwise comparison accuracy. For example, on GenAI-Bench, \score (reg) achieves $78.5$ accuracy, which is higher than the $59.0$ of the \score (gen). The results are similar for the other benchmarks. We thus conclude that regression scoring with more fine-grained scores is a better choice.

\section{Conclusion}
\label{sec:conclusion}
In this paper, we introduce \score, which is trained on our meticulously curated dataset \data for video evaluation and can serve as good simulator for human feedback on generated videos. We hired 20 expert raters to annotate the 37.6K videos generated from 11 popular text-to-video generative models across 5 key aspects, Visual Quality, Temporal Consistency, Dynamic Degree, Text-to-Video Alignment and Factual Consistency. Our IAA match ratio gets more than 60\%. 
We test the performance of \score using Spearman correlation on \data-test and EvalCrafter, and using pairwise comparison accuracy on GenAI-Bench and VBench. The results show that \score consistently gets the best performance, surpassing the powerful baseline GPT-4o and Gemini 1.5 Flash/Pro by a large margin. 
Our work highlights the importance of using MLLM for video evaluation and demonstrates the future prospects of simulating human scores or feedback in generative tasks, due to its rich world knowledge and the high-quality rating dataset with fine-grained and multiple dimensions.

\section*{Acknowledgement}
We express our gratitude to StarDust for providing video raters and to DataCurve for supplying the GPU compute resources. Additionally, we express our thanks to all the raters who offered valuable feedback and suggestions, which were instrumental in completing this work.

\bibliography{acl_latex}

\appendix
\clearpage
\label{sec:appendix}

\section{Ethical Statement}
This work fully complies with the ACL Ethics Policy. We declare that there are no ethical issues in this paper, to the best of our knowledge.

\section{Risks and Limitation}
\label{sec:limitation}
Although we have designed systematic pipelines to recruit expert raters and annotate the video evaluation scores, we still find out that some annotations contain errors and may harm the overall quality of the dataset. Our IAA score computation is only based on a small number of trial examples and, thus might not represent the actual IAA of the whole annotations. Besides, while \score is proven to be able to effectively give reasonable scores on our defined five aspects, it can still sometimes output wrong scores that do not match our expectations. We admit this drawback and list that as one of our future works.

\section{Dataset Licence}
We have used VidProM~\cite{Wang2024VidProMAM} to collect the prompts used for video generation, whose usage LICENSE is CC BY-NC 4.0 license. For other evaluation datasets, We did not find license for EvalCrafter~\cite{liu2023evalcrafter} human annotations. GenAI-Bench~\cite{jiang2024genai} is under MIT licence, and VBench~\cite{Huang2023VBenchCB} is under Apache 2.0 license. We are thus able to utilize these datasets in our experiments.

We also release our curated dataset, \data, under MIT license to contribute to the video evaluation dataset.

\section{Annotator Management}
During the annotation, we have recruited 20 expert raters, where 14 of them are undergraduate or graduate students, who will become one of the authors of our paper, and the rest of them are assured to be paid with decent salary.

\section{Video Format Normalizing Details}
\label{subsec:video_format_norm}
To mitigate difference of videos format from different generative models, we normalize the frame rate of all the generated videos to 8 fps (frames per second). Specifically, for high frame rate model Pika and AnimateDiffusion~\cite{guo2023animatediff}, we use uniform down-sampling to normalize Pika from 24 fps to 8fps, and AnimateDiffusion from 23 fps to 8 fps. For low frame rate model Text2Video-Zero~\cite{text2video-zero}, we use video frame interpolation model RIFE~\cite{huang2022realtime} to interpolate frames, adding the frame rate from 4 fps
 to 8 fps. For real-world videos from DiDeMo~\cite{hendricks2017localizing} and Panda70M~\cite{chen2024panda70m} in post augmentation of \data, we use the same down-sampling as Pika and AnimateDiffusion to reduce their frame rate from 30 fps to 8 fps.

 Additionally, since video from Pika are always attached a watermark "PIKA-LABS", we cropped all the Pika videos from the resolution of (1088, 640) to (768, 480), making Pika video indistinguishable from videos from other models.


\begin{figure*}
    \centering
    \includegraphics[scale=0.5]{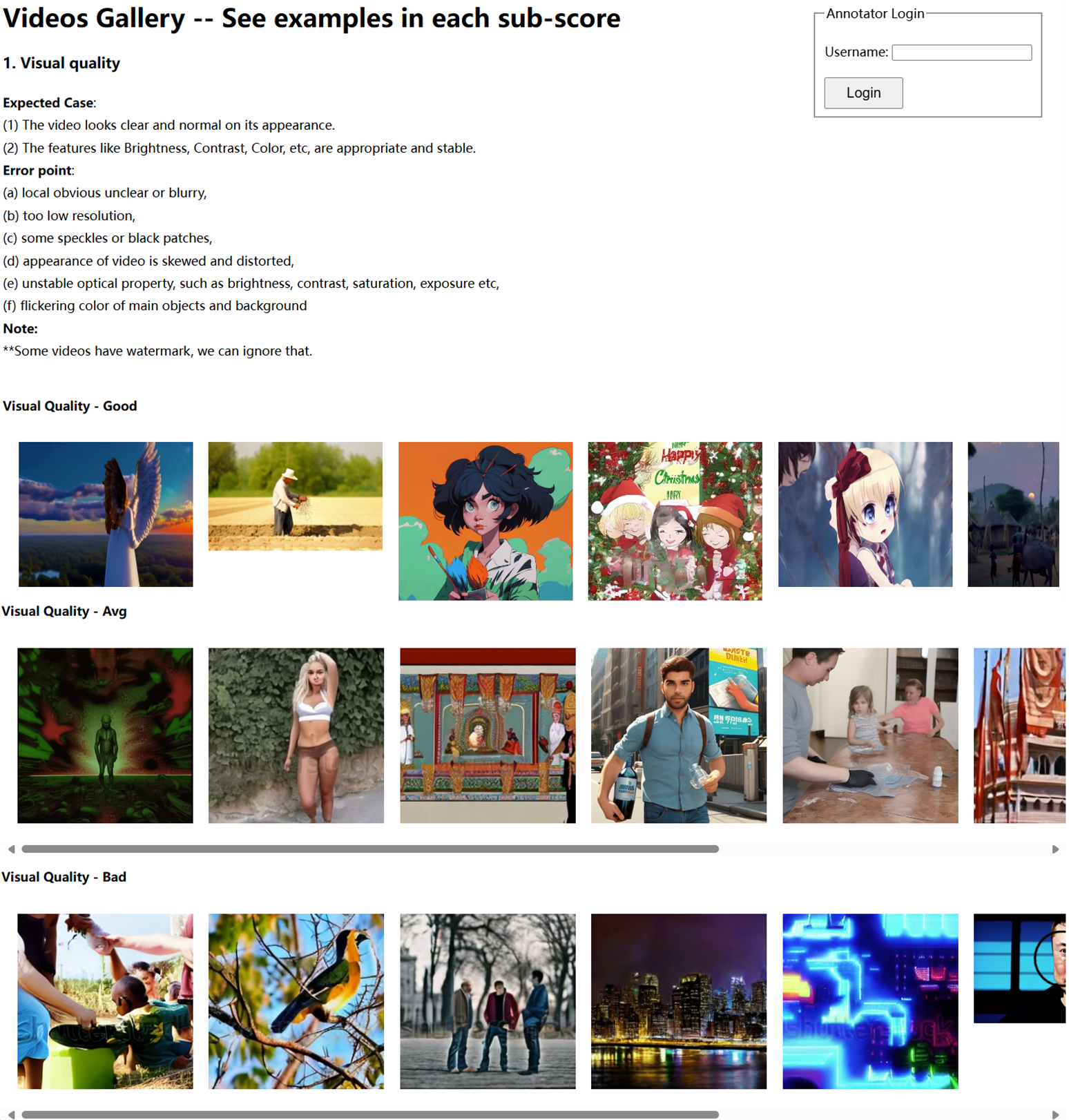}
    \caption{Welcome Page of our video annotating website, with definition, checklist for error points and diverse video examples.}
    \label{fig:anno_webpage_welcome_new}
\end{figure*}


\begin{figure*}[!ht]
    \centering
    \includegraphics[width=\textwidth]{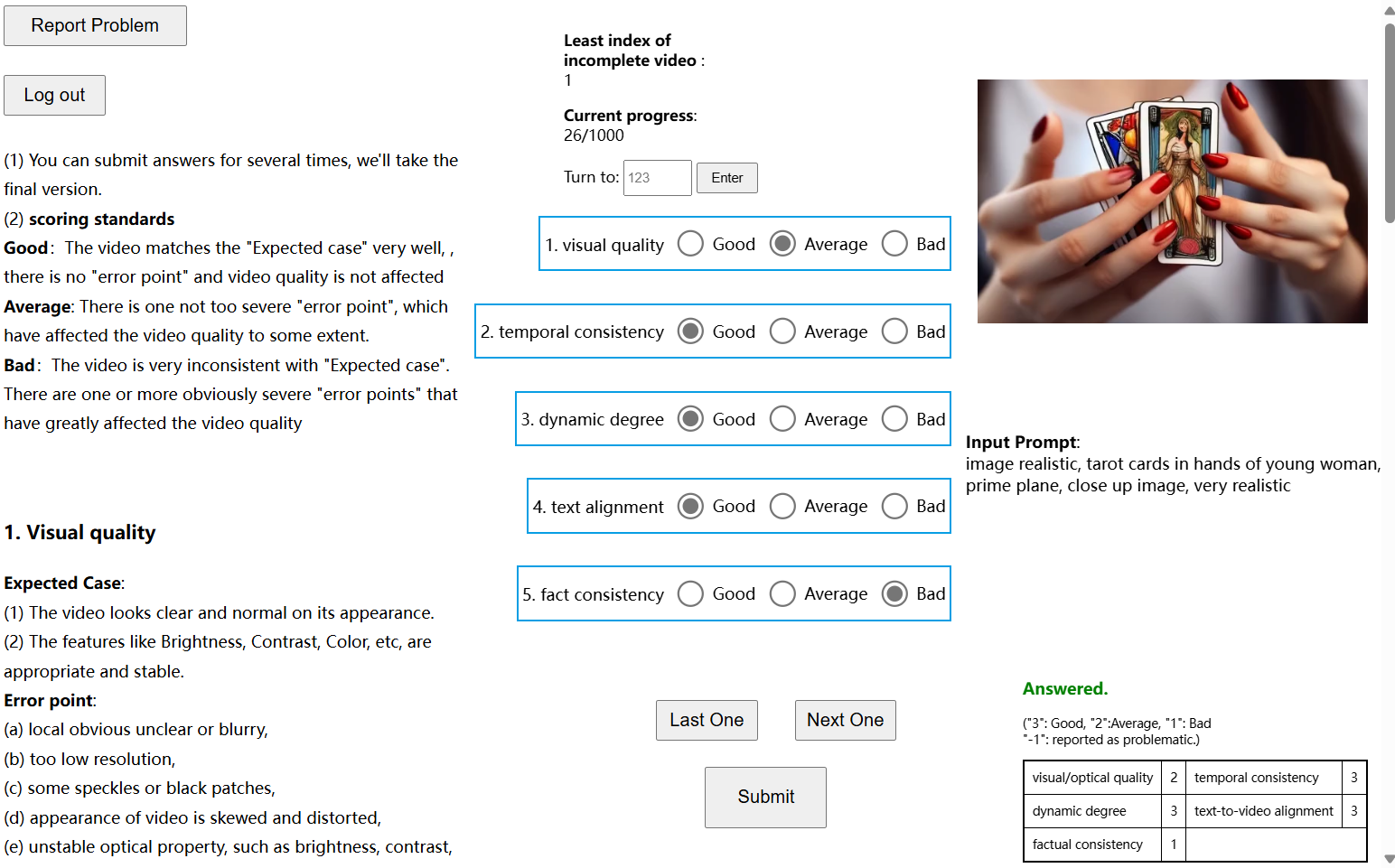}
    \caption{Working page of our video annotating website}
    \label{fig:anno_webpage_main}
\end{figure*}

\section{Annotation Details}
Additional annotation details are put in this section for the reference.

Firstly we show the user interface of our annotating website in ~\autoref{fig:anno_webpage_welcome_new} and ~\autoref{fig:anno_webpage_main}. In both welcome page and working page, we list the definition and a checklist of error points in five evaluation dimensions, as shown in ~\autoref{tab:aspect_error_points}. Additionally we also provide many Good/Average/Poor videos as examples in each dimension for raters to quickly understand each dimension and align well with our understanding. 
\begin{table*}[!ht]
\centering
\small
\begin{tabular}{p{1.5cm}|p{14cm}}
\toprule
Evaluation Aspect & Detailed Description for Annotation \\
\midrule
Visual Quality & \begin{tabular}{p{0.85\textwidth}}
Expected Case: \\(1) The video looks clear and normal on its appearance.\\ (2) The features like Brightness, Contrast, Color, etc, are appropriate and stable.\\ Error point: \\(a) local obvious unclear or blurry, (b) too low resolution,  (c) some speckles or black patches, (d) appearance of video is skewed and distorted, (e) unstable optical property, such as brightness, contrast, saturation, exposure etc, (f) flickering color of main objects and background\\ Note:Some videos have watermark, we can ignore that.\end{tabular} \\
\midrule
Temporal Consistency & \begin{tabular}{p{0.85\textwidth}}
Expected Case:\\ (1) The main objects, main characters and overall appearance are consistent \\ across the video.\\ (2) The appearance of video as well as the movements of humans and objects \\ are smooth and natural.\\ Error points:\\ (a) The person or object suddenly disappears or appears, (b) The type or class of objects has obvious changes, (c) There is an obvious switch in the screen shot,(d) the appearance of video or movements in it is laggy and un-smooth, (e) local deformation or dislocation of human or objects due to the motion.\\ (for large scale deformation, the video should also be rated as bad in "1. visual quality").\\ Note:\\ For a video almost static or with small dynamic degree, as long as it does not have error points, then it should be scored as good.\end{tabular} \\
\midrule
Dynamic Degree & \begin{tabular}{p{0.85\textwidth}}
Expected Case:\\ (1) The video is obviously not static, the people or objects or the video screen \\ is dynamic.\\ (2) The video can be easily distinguished from a static image.\\ Note:\\ You are supposed to focus on only dynamic degree, regardless of the visual quality and video content\end{tabular} \\
\midrule
Text-to-Video Alignment & \begin{tabular}{p{0.85\textwidth}}
Expected Case:\\ The characters, objects, motions, events etc. that are mentioned in text input prompts all exist reasonably.\\ Error points:\\ (a) The people and objects in prompt do not appear in video, (b) The actions and events in prompt do not appear in video, (c) The number, size, shape, color, state, movement and other attributes of the objects in the video do not match the prompt, (d) Text mentioned in prompt is not displayed correctly in the video, such as "a placard saying 'No Smoking'" but "No Smoking" is not spelled correctly in the video, (e) The video format (such as width, height, screen ratio, duration) does not match the format in prompt.\end{tabular} \\

\midrule

Factual Consistency & \begin{tabular}{p{0.85\textwidth}}
Expected Case:\\ (1) Overall appreance and motion are consistent with our common-sense, \\ physical principles, moral standards, etc.\\ Error points:\\ (a) static ones: Content in video goes against common sense in life, such as \\ lighting a torch in the water, standing in the rain but not getting wet, etc.\\ (b) static ones: The size, color, shape and other basic properties of objects violate scientific principles\\ (c) dynamic ones: The overall movement of people or objects violates common-sense and laws of physics, such as spontaneous upward movement against gravity, abnormal water flow, etc.\\ (d) dynamic ones: Partial movements of people or objects violate common-sense and laws of physics, such as the movement of hands or legs is anti-joint, etc.\\ Notes:\\ Relation with '5. text-to-video alignment':\\ Some text prompts express fictional and unrealistic content, for example, "a dog plays the guitar in the sky" or "an astronaut rides a horse in space".  In this case, regardless of the veracity of the text prompt, you should \\ only consider whether the other content in the video makes sense.\end{tabular} \\
\bottomrule
\end{tabular}
\caption{Expected cases and error cases for each aspect that annotators can see during the annotation.}
\label{tab:aspect_error_points}
\end{table*}

\section{Prompting Template}
In process of training Mantis~\cite{jiang2024mantis} for generation scoring and the testing with "MLLM Prompting" baselines, we use the same prompt template provided in~\autoref{tab:mllm_prompt_template_gen}. 

For training Mantis with regression scoring, we make modification to the above template accordingly, instructing model to output a float number ranges from 1.0 to 4.0, as shown in ~\autoref{tab:mllm_prompt_template_reg}.

\begin{table*}[!ht]
    
    \small
    \centering
    \begin{tabular}{l}
\toprule
Suppose you are an expert in judging and evaluating the quality of AI-generated videos, \\
please watch the following frames of a given video and see the text prompt for generating the video, \\
then give scores from 5 different dimensions: \\
(1) visual quality: the quality of the video in terms of clearness, resolution, brightness, and color \\
(2) temporal consistency, the consistency of objects or humans in video \\
(3) dynamic degree, the degree of dynamic changes \\
(4) text-to-video alignment, the alignment between the text prompt and the video content \\
(5) factual consistency, the consistency of the video content with the common-sense and factual knowledge \\
\\
For each dimension, output a number from [1,2,3,4], \\
in which '1' means 'Bad', '2' means 'Average', '3' means 'Good', \\
'4' means 'Real' or 'Perfect' (the video is like a real video) \\
Here is an output example: \\
visual quality: 4 \\
temporal consistency: 4 \\
dynamic degree: 3 \\
text-to-video alignment: 1 \\
factual consistency: 2 \\
\\
For this video, the text prompt is \textbf{"\{text\_prompt\}"}, \\
all the frames of video are as follows: \\
\bottomrule
    \end{tabular}
\caption{Prompting template in generation format used for \score training and the MLLM prompting baselines}
\label{tab:mllm_prompt_template_gen}
\end{table*}
\begin{table*}[!ht]
    
    \small
    \centering
    \begin{tabular}{l}
\toprule
Suppose you are an expert in judging and evaluating the quality of AI-generated videos, \\
please watch the following frames of a given video and see the text prompt for generating the video, \\
then give scores from 5 different dimensions: \\
(1) visual quality: the quality of the video in terms of clearness, resolution, brightness, and color \\
(2) temporal consistency, the consistency of objects or humans in video \\
(3) dynamic degree, the degree of dynamic changes \\
(4) text-to-video alignment, the alignment between the text prompt and the video content \\
(5) factual consistency, the consistency of the video content with the common-sense and factual knowledge \\
\\
For each dimension, output a float number from 1.0 to 4.0, \\
higher the number is, better the video performs in that dimension, \\
the lowest 1.0 means Bad, the highest 4.0 means Perfect/Real (the video is like a real video) \\
Here is an output example: \\
visual quality: 2.24 \\
temporal consistency: 3.89 \\
dynamic degree: 3.17 \\
text-to-video alignment: 1.86 \\
factual consistency: 2.16 \\
\\
For this video, the text prompt is \textbf{"\{text\_prompt\}"}, \\
all the frames of video are as follows: \\
\bottomrule
    \end{tabular}
\caption{Prompting template used for the MLLM prompting baseline and \score training}
\label{tab:mllm_prompt_template_reg}
\end{table*}

\section{Feature-based Baselines Discretization}
As described in ~\autoref{baselines}, we employ several statistical or neural feature-based metrics as baselines for comparison with our model. The continuous float-format outputs of these metrics are discretized into labels [1, 2, 3, 4], aligning with our annotation data format. The discretization rules are presented in ~\autoref{tab:dis_rules}. Metrics with a $\uparrow$ symbol indicate that higher values are better, while those with a $\downarrow$ symbol indicate that lower values are better.

\begin{table*}[!ht]

\small
\centering
\begin{tabular}{c|c|cccc}
\toprule
Dimension &Metric  & 1 (Bad) & 2 (Avg)  & 3 (Good) & 4 (Perfect)    \\
\midrule
\multirow{2}{*}{Visual Quality} & PIQE$\downarrow$ & $[50,\infty)$ & [30,50) & [15,30) &  [0,15)  \\
      & BRISQUE$\downarrow$ & [50,$\infty$) & [30,50) & [10,30) &  [0,10)  \\
\midrule
\multirow{3}{*}{Temporal Consistency} & CLIP-sim$\uparrow$ & [0,0.80) & [0.80,0.90) & [0.90,0.97) &  [0.97,1]  \\
 & DINO-sim$\uparrow$ & [0,0.75) & [0.75,0.85) & [0.85,0.95) &  [0.95,1]  \\
 & SSIM-sim$\uparrow$ & [0,0.6) & [0.6,0.75) & [0.75,0.9) &  [0.9,1]  \\
\midrule
\multirow{2}{*}{Dynamic Degree} & MSE-dyn$\uparrow$ & [0,100) & [100,1000) & [1000,3000) &  [3000,$\infty$)  \\
 & SSIM-dyn$\downarrow$ & [0.9,1] & [0.7,0.9) & [0.5,0.7) &  [0,0.5)  \\
\midrule
\multirow{2}{*}{Text-to-Video Alignment} & CLIP-Score$\uparrow$ & [0.2,0.27) & [0.27,0.31) & [0.31,0.35) &  [0.35,0.4]  \\
 & X-CLIP-Score$\uparrow$ & [0,0.15) & [0.15,0.23) & [0.23,0.30) &  [0.30,1]  \\

\bottomrule
\end{tabular}
\caption{Discretization rules for featured-based baselines.}
\label{tab:dis_rules}
\end{table*}

\section{Case study of \data}
\label{sec:case_study}

\begin{figure}[!ht]
\centering
        \includegraphics[scale=0.28]{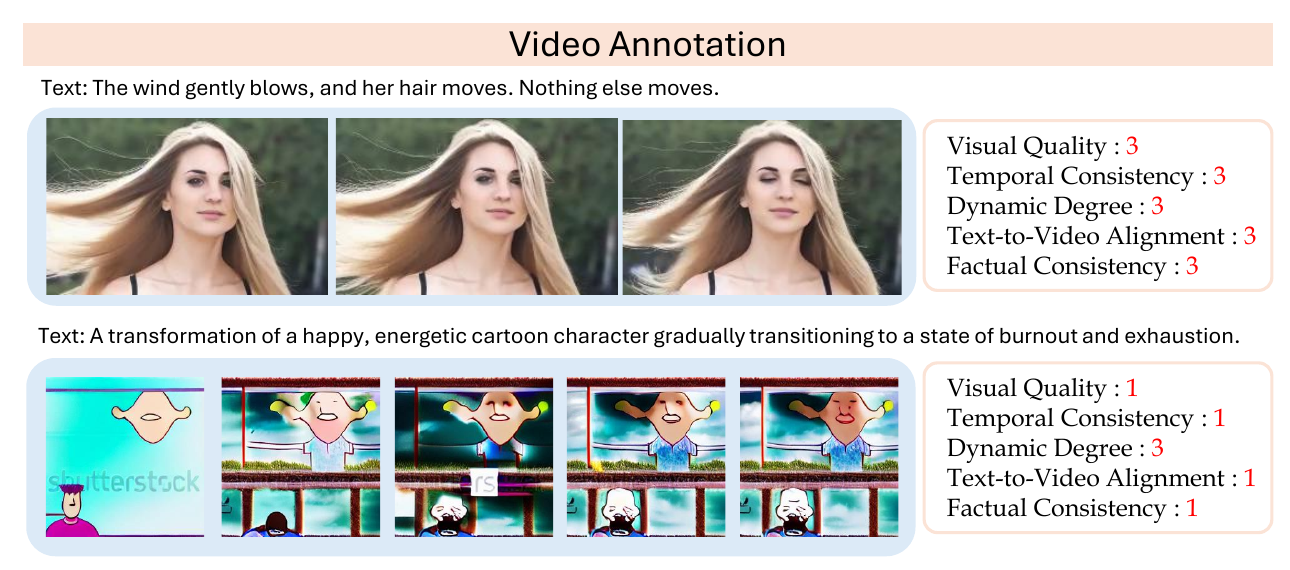}
        \captionof{figure}{Example of annotations. Each video has a text description and is rated for the 5 aspects.}
        \label{fig:case_study}
\end{figure}

\begin{figure}[!ht]
        \centering
        \includegraphics[scale=0.32]{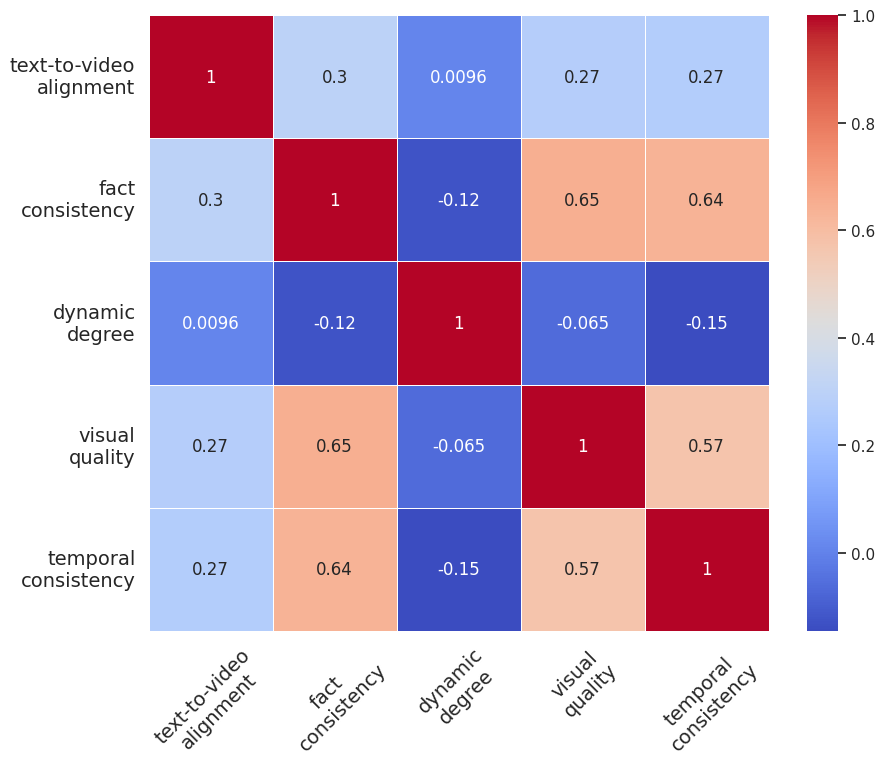}
        \caption{Correlation study on the evaluation aspects.}
        \label{fig:correlation}
\end{figure}

We showcase the annotations examples in~\autoref{fig:case_study}. The first example depicts a clear video of a woman with her hair moving, thus scoring 3 in all 5 aspects. The second example shows a distorted video, thus scoring 1 across all the aspects except the dynamic degree. We further analyzed the correlations between the designed aspects in~\autoref{fig:correlation}. We found that visual quality achieves a high correlation of 0.6 with temporal consistency, while dynamic degree has a very low correlation with all other aspects.

\section{Leaderboard}
We generated 200 videos using various T2V models with prompts sampled from our prompt set, then we rank these T2V models based on the average score of five dimensions in \score, as shown in ~\autoref{tab:leaderboard}.

\begin{table*}[!ht]
\renewcommand{\arraystretch}{1.3}
\small
\centering
\begin{tabular}{l|cccccc}

\toprule
T2V Model  & Avg &  VQ & TC & DD & TVA & FC    \\
\midrule
Kling              & \textbf{81.83} & \textbf{84.63} & \textbf{82.49} & \textbf{85.10}  & \textbf{74.29} & \textbf{82.65} \\
OpenSora-v1.2      & 69.85 & 70.36 & 67.17 & 75.91 & 68.17 & 67.62 \\
Morph Studio       & 67.56 & 66.79 & 70.98 & 66.71 & 68.55 & 64.77 \\
Morph Studio       & 67.56 & 66.79 & 70.98 & 66.71 & 68.55 & 64.77 \\
VideoCrafter-2     & 66.23 & 66.72 & 68.81 & 67.04 & 65.77 & 62.79 \\
Gen-2              & 65.74 & 65.14 & 67.37 & 67.97 & 65.39 & 62.84 \\
PikaLab            & 65.72 & 67.21 & 67.98 & 65.17 & 63.79 & 64.43 \\
Video-LaVIT        & 64.04 & 67.57 & 59.68 & 70.83 & 66.7  & 55.40  \\
LaVie-base         & 62.86 & 63.11 & 59.33 & 70.25 & 62.40  & 59.20  \\
MagicTime          & 62.61 & 65.05 & 63.2  & 67.33 & 61.99 & 55.50  \\
HotShot-XL         & 62.31 & 58.55 & 59.63 & 70.90  & 63.93 & 58.54 \\
Latte              & 62.02 & 63.34 & 59.23 & 68.38 & 62.56 & 56.60  \\
OpenSora-v1.0      & 61.94 & 62.49 & 58.75 & 69.65 & 64.30  & 54.51 \\
OpenSora-v1.1      & 61.29 & 61.50  & 56.82 & 69.82 & 64.72 & 53.61 \\
VideoCrafter-1-512 & 60.92 & 60.94 & 60.12 & 67.38 & 60.44 & 55.73 \\
AnimateDiff        & 57.33 & 64.83 & 41.89 & 73.46 & 65.76 & 40.69 \\
ModelScope         & 52.45 & 47.41 & 49.76 & 69.14 & 54    & 41.94 \\
LVDM               & 47.20  & 33.15 & 44.01 & 72.75 & 58.85 & 27.25 \\
ZeroScope\_576w    & 44.69 & 31.35 & 39.65 & 73.76 & 49.40  & 29.27 \\

\bottomrule
\end{tabular}
\caption{Leaderboard of \score for existing text-to-video models on 200 curated examples.}
\label{tab:leaderboard}
\end{table*}



\end{document}